\documentclass[11pt,a4paper]{article}
\usepackage[hyperref]{acl2020}
\usepackage{times}
\usepackage{latexsym}
\usepackage[T1]{fontenc}
\usepackage{booktabs}
\usepackage{longtable}
\usepackage{graphicx}

\usepackage{microtype}

\aclfinalcopy 

\title{UniCase - Rethinking Casing in Language Models}

\author{Rafał Powalski \\
  Applica.ai  \\
  \texttt{rafal.powalski@applica.ai} \\\And
  Tomasz Stanisławek \\
  Applica.ai \\
  Warsaw University of Technology \\
  \texttt{tomasz.stanislawek@applica.ai} \\}

\date{}

\begin{document}
\maketitle
\begin{abstract}

In this paper, we introduce a new approach to dealing with the problem of case-sensitiveness in Language Modelling (LM). We propose simple architecture modification to the RoBERTa language model, accompanied by a new tokenization strategy, which we named Unified Case LM (UniCase). We tested our solution on the GLUE benchmark, which led to increased performance by 0.42 points. Moreover, we prove that the UniCase model works much better when we have to deal with text data, where all tokens are uppercased (+5.88 point).\footnote{Work in progress.}

\end{abstract}

\section{Introduction}

Many natural languages in their written form encode some information in the case of the letter: the beginning of the sentence, proper nouns, headings of publication titles, to name a few. People can process that kind of information in a special way, learning word semantics separately from the case information. However, state-of-the-art (SOTA) approaches for building Language Models do not use this property \cite{devlin2019bert,liu2019roberta,raffel2020exploring,brown2020language}. As an example, consider RoBERTa language model, where we have different tokenizer outputs for each word case type (lower, title, upper) and multiple vocabulary entries for the same word but different cases (see Table~\ref{tab:tokenization-example}). 

Whereas different approaches to dealing with case-sensitivity issues were proposed and tested on Machine Translation systems \cite{etchegoyhen-gete-2020-case, case_sensitive_lm_10.1007/978-3-030-47426-3_51} where the impact of the casing can be significant, there are no similar studies in the context of building Language Models. In fact, most of LMs use cased (with the original text) and uncased (with lowercased text) version of the model \cite{devlin2019bert,liu2019roberta,xlm-robertaconneau2020unsupervised}. 

In this paper, we present a model that resolves the problems presented in Table~\ref{tab:tokenization-example}. Particularly, we provide two main contributions. Firstly, we propose UniCase: a new language model based on transformer architecture with novel tokenization strategy. Our method improves RoBERTa scores by 0.42 on the GLUE benchmark~\cite{wang-etal-2018-glue}. Secondly, we test our new UniCase model on noisy texts, where true case is unknown (all texts were converted to uppercase or lowercase). We will release all pretrained models in an open, public repository.

\begin{table}
\centering
\footnotesize{
\begin{tabular}{lrl}
\hline \textbf{Word} & \textbf{Vocab subtokens} \tiny{(\_ means spaces)} \\ \hline
acknowledgement & \_acknowledgement \\
Acknowledgement & \_A/cknowled/gement \\
ACKNOWLEDGEMENT & \_AC/KN/OW/LED/G/EMENT \\
\hline
other & \_other \\
Other & \_Other \\
OTHER & \_OTHER \\
\hline
\end{tabular}
}
\caption{\label{tab:tokenization-example} RoBERTa tokenization examples. \textit{Acknowledgement} is tokenized differently for each  case variant and \textit{other} has redundant entries in the vocabulary.}
\end{table}

\section{Related work}

\subsection{Language Modelling}

State-of-the-art approaches \cite{devlin2019bert,liu2019roberta,raffel2020exploring,brown2020language} for building Language Models use Transformer architecture \cite{vaswani2017attention} with BPE~\cite{bpe_sennrich-etal-2016-neural} or Unigram LM-based tokenization methods~\cite{unigram_kudo2018subword}, where each subtoken from the vocabulary has only one semantic embedding in the model. With that architecture there are two common approaches for dealing with the problem of case-sensitiveness in Language Modelling: cased (with original text)~\cite{liu2019roberta,xlm-robertaconneau2020unsupervised,raffel2020exploring,sun2019ernie20} and uncased (with lowercased text)~\cite{devlin2019bert,iandola2020squeezebert,sanh2020distilbert}, where cased models proved to be more suitable for majority of NLP tasks~\cite{wang-etal-2018-glue,wang2020superglue,devlin2019bert}. 

\subsection{Tokenization} 
\label{sec:tokenization_methods}

Tokenization is a way of splitting a text into tokens, which NLP models use as smallest piece of information. Over the years researchers have been introducing different approaches to tokenization  with three types of tokens as bases: words, characters and subwords. Subwords are considered to be the most effective one~\cite{study_tokenization_10.1007/978-981-15-6198-6_18}. 

Byte-Pair-Encoding (BPE)~\cite{bpe_sennrich-etal-2016-neural} segmentation balances vocabulary size and the length of a sequence proccessed by the model in the single pass. The general idea behind BPE is to create the vocabulary by iteratively merging the most frequent pair of characters or subtokens into a new one. Thus, words with low frequencies in the corpus will be represented as combinations of multiple subtokens or characters. It turns out, that this solution has its own drawbacks, such as lack of multiple segmentations endowed with probabilities, and regularization techniques. These two issues were fixed by introducing tokenization based on unigram language model, which can produce  multiple subword segmentations endowed with probabilities~\cite{unigram_kudo2018subword}. It has been proved that language models based on Unigram LM tokenizer work better~\cite{bpe_suboptimal_bostrom2020byte}.

\subsection{Neural Machine Translation} 

Neural Machine Translation (NMT) has recently been the subfield of NLP where many new tokenization techniques were introduced, before being more widely adopted by the whole NLP community~\cite{bpe_sennrich-etal-2016-neural,unigram_kudo2018subword}. This is also true when it comes to the new approaches to encoding case information into the neural models~\cite{berard2019naver,etchegoyhen-gete-2020-case,case_sensitive_lm_10.1007/978-3-030-47426-3_51}. The following methods are commonly used in NMT: \textbf{Truecasing}, \textbf{Recasing}, \textbf{Case Factors (CF)} and \textbf{Inline Casing (IC)}. Two of them can be naturally adopted to the problem of language modelling: CF (subtoken embedding is the concatenation of lowercased base-token embedding and the case embedding for each variant i.e. title, uppercase, mixed) and IC (working on lowercased text but adding extra tags before words which indicate case variants). In this paper we propose a solution which is similar to the Case Factors method with some modifications.

\section{Method}

In this section we describe our approach to dealing with casing in language models. It can be decomposed into two main parts: tokenization and model architecture.

\subsection{Tokenization}

As described in Section \ref{sec:tokenization_methods} Unigram tokenizer outperforms BPE on Language Model pretraining \cite{bpe_suboptimal_bostrom2020byte}, therefore we are basing our solution on this method. In addition, we want to create a tokenizer which is able to fulfil the following conditions:
\begin{itemize}
    \item tokenization should be the same for texts regardless of different casing, with the exception of the next point 
    \item the above condition can be violated for words written with mixed casing (e.g. camelCasing), such words could be split in places where letter casing changes (e.g. word \textit{RoBERTa} could be splitted into \textit{\_Ro/BERT/a}, even though word \textit{Roberta} could be represented as a single token).
\end{itemize}

In order to do that, we trained the Unigram Sentencepiece tokenizer~\cite{sentencepiece_kudo-richardson-2018} in a way, where word tokens are kept in various case variants. We obtained that, in order to satisfy above conditions, we need only 3 case variants (shapes) i.e. lowercase (aaa), uppercase (AAA) and titlecase (Aaa). Other shapes are only needed in mixed case variants which we decided to split. It is worth mentioning, that such token multiplication by its shape variants is valid only for word tokens. For tokens which contain numbers, punctuation marks, etc. we kept only the original.


\subsection{Model}

\begin{figure*}[t]
  \centering
  \includegraphics[width=0.8\textwidth]{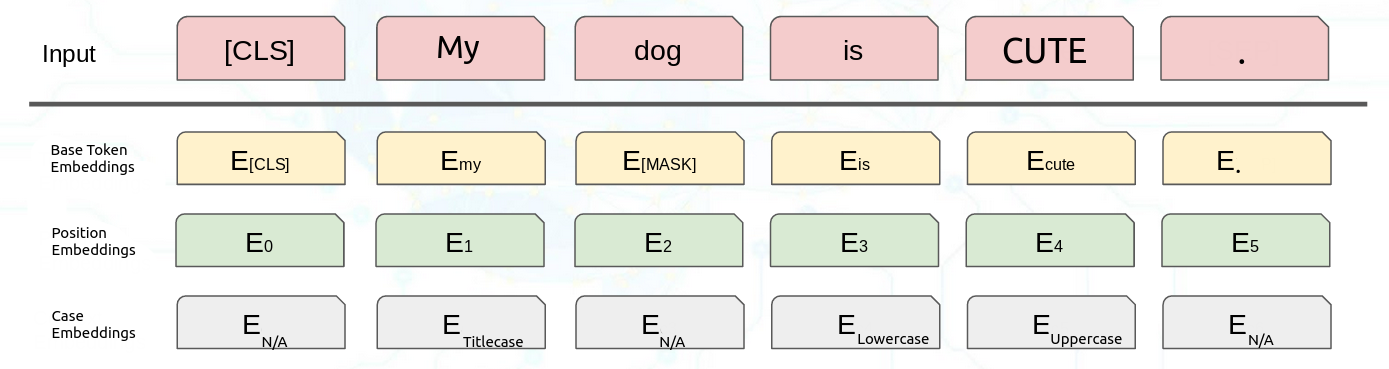}
  \caption{Unicase input representation. The input embeddings are the sum of the base token embeddings, the positional embeddings and case embeddings.}
  \label{fig:caseemb}
\end{figure*}

Model architecture is modified in order to utilize the information, that some word tokens in the dictionary, are linked with the same base (lowercase) form. To this end, we decompose the token embedding into base-token embedding and case embedding (Figure \ref{fig:caseemb}). These embeddings are trainable and added to each other in the same fashion as positional embeddings are added in the original Transformer architecture \cite{vaswani2017attention}. In the consequence of such decomposition, models with the same number of parameters, can utilize much bigger vocabularies.


During the pretraining phase, we are also decomposing the original masked token prediction task into base-token prediction and case prediction tasks. Final loss is computed as a weighted sum of two tasks' losses (\ref{eq:loss_sum}). By using weights, we are forcing the model to focus more on base-token prediction. On the initial setting we chose $\alpha=0.1$.

\begin{equation}
  \label{eq:loss_sum}
  L = L_{base\_token} + \alpha~L_{case}
\end{equation}

\section{Experiments}


\subsection{Implementation}

We are basing our UniCase architecture on RoBERTa code implemented in FAIRSEQ~\cite{ott2019fairseq}. Code and pretrained models will be publicly released.

\subsection{Unsupervised training}

\subsubsection{Models}
\label{sec:unicase-models}

We have trained two versions of UniCase model corresponding to different tokenizer settings.

\begin{itemize}
    \item UniCase model based on UniCase Tokenizer with $20k$ base tokens, which correspond to $base\_token\_embedding\_size=20k$ and $vocab\_size \approx 57k$
    \item UniCase model based on UniCase Tokenizer with $32k$ base tokens, which correspond to $base\_token\_embedding\_size=32k$ and $vocab\_size \approx 90k$
\end{itemize}
Both models correspond to RoBERTa-base in terms of size. 

As a baseline model we chose RoBERTa-base architecture and trained it from scratch. We modified original setup and used a Unigram tokenizer ($vocab\_size=32k$) to be sure that potential performance deterioration is not caused by the BPE tokenizer, which was proved to be not an optimal choice for tokenization~\cite{bpe_suboptimal_bostrom2020byte}.

We have trained all models on DGX-2 server using the setting recommended by RoBERTa authors with $batch\_size=2048$ for $100k$ update steps.

\subsubsection{Data}

The size and quality of pretraining data was proved to play important role for achieving state-of-the-art results~\cite{liu2019roberta,brown2020language}. Thus, all our models were pretrained on the CCNet dataset~\cite{wenzek2019ccnet}, which contains about 700 millions of documents for English language, corresponding to 330 GB of uncompressed text.

\subsection{Experiments}

\begin{table*}[t]
\centering
\begin{tabular}{@{}llllllllll@{}}
\toprule
Model & CoLA & MNLI & MRPC  & QNLI   & QQP & RTE & SST & STS-B  & Average \\ \midrule
\multicolumn{10}{c}{Original casing} \\
\midrule
UniCase (20k) & \textbf{59.82} & \textbf{85.40} & \textbf{90.89}  & 91.22  & 88.01  & 69.68  & 92.83  & 87.96  & 83.20  \\
UniCase (32k) & 58.29  & 85.18  & 90.88  & \textbf{91.29} & \textbf{88.16} & \textbf{71.84} & 92.78  & \textbf{88.18}  & \textbf{83.29} \\
RoBERTa (32k) & 57.92 & 84.84  & 90.33  & 91.01  & 88.14  & 69.31  & \textbf{94.04} & 87.80  & 82.87  \\
\midrule
\multicolumn{10}{c}{All texts from train and development sets were lowercased}\\
\midrule
UniCase (32k) & \textbf{55.99} & \textbf{85.23} & \textbf{90.85} & \textbf{90.90} &	88.13 &	\textbf{70.40} & 92.89 & \textbf{88.26} & \textbf{82.80} \\
RoBERTa (32k) & 55.08 &	84.82 &	90.65 &	90.31 &	\textbf{88.14} & 67.87 & \textbf{94.15} &	87.70 &	82.31 \\ 
\midrule
\multicolumn{10}{c}{All texts from train and development sets were uppercased}\\
\midrule
UniCase (32k) & \textbf{56.25} & \textbf{85.19} & \textbf{91.28} & \textbf{91.10} &	\textbf{88.09} & \textbf{71.84} & \textbf{92.83} & \textbf{88.11} &	\textbf{83.07} \\
RoBERTa (32k) & 39.24 &	80.11 &	87.84 &	87.23 &	86.90 &	62.82 &	89.11 &	85.21 &	77.19 \\ 
\bottomrule
\end{tabular}
\caption{Results on GLUE. The “Average” column is slightly different than the official GLUE score, since we exclude the problematic WNLI set. F1 scores are reported for QQP and MRPC, Spearman correlations are reported for STS-B, and accuracy scores are reported for the other tasks. All task results are median over four runs.}
\label{tab:results-table} 
\end{table*}

\subsubsection{Data}

We conducted all our experiments on the GLUE benchmark~\cite{wang-etal-2018-glue}, which is a collection of well known datasets for testing natural language understanding systems. The original benchmark contains 9 tasks, from which we skip the problematic WNLI set. All our results were based on development sets.

\subsubsection{Settings}

We have trained all models separately for each of the GLUE tasks, using only the training data for the corresponding task. For finetuning on each task we have used parameters recommended by RoBERTa authors~\cite{liu2019roberta}. All results presented in tables are medians over four random initializations.

\subsubsection{Results on original texts}

At the beginning, we evaluated all pretrained models (described in section~\ref{sec:unicase-models}) by using text in original casing (see Table~\ref{tab:results-table}. We observed that both UniCase models variants perform better than baseline model. Only on SST dataset RoBERTa model is better. Interestingly, this is the only task in GLUE where text is lowercased\footnote{Further investigation is needed to determine the reason of that.}. The influence of vocabulary size in UniCase model is not exactly known as more experiments need to be done . Nevertheless, the best model was trained with vocabulary size containing 32k of base-tokens. Therefore, that model will be used in further experiments.

\subsubsection{Results on noisy texts}

Firstly, intuitively all models achieved better scores with original casing but only UniCase model have stable overall results between different experiment settings: original (83.29), lowercase (82.80) and uppercase (83.07). Surprisingly, on MRPC dataset, UniCase model achieved higher score on average where texts was uppercased, which might be explained with very small development set size and higher deviation of results.

We find that overall performance of the UniCase model is better that RoBERTa when we must deal with noisy texts~\ref{tab:results-table}. However, only in the case where texts where uppercased, we can observe significant difference between these two models, which is 5.88 points based on average metric from presented GLUE tasks. We think that tokenization of uppercase document is heavily fragmented which in conjunction with relatively poor uppercase representation in pretraining corpora might lead to poor uppercase text understanding. 


\section{Conclusion}

We have presented and tested new UniCase architecture dealing with case-sensitivity in Language Modelling by decomposing information about casing into a separate component. Consequently, we were able to build models with the same number of parameters utilizing larger vocabularies. In contrast to classic Language Models, UniCase does not have to build a semantic understanding of words or sentences in all case variants, potentially leading to more effective training. 

We showed that our method outperforms the RoBERTa baseline on almost all tested tasks. Notably, results reported on uppercased GLUE tasks show that models trained with our method understand uppercased documents much better. That shows a promising application of UniCase models in understanding documents where uppercasesed letters or words are more common, i.e., business documents, forms, invoices.

\section*{Acknowledgments}

We thank Filip Graliński, Łukasz Garncarek, Michał Pietruszka, Łukasz Borchmann and Dawid Jurkiewicz for their discussion about the paper and ours managing directors at Applica.ai: Adam Dancewicz and Piotr Surma.

The authors would like to acknowledge the support the Applica.ai project has received as being co-financed by the European Regional Development Fund (POIR.01.01.01-00- 0144/17-00). 

\bibliography{paper_bib}
\bibliographystyle{acl_natbib}

\end{document}